\documentclass[letterpaper, 10 pt, conference]{ieeeconf}  

\IEEEoverridecommandlockouts                              

\usepackage{tabularx,booktabs}
\usepackage{amsmath,amssymb,amsfonts}
\usepackage{graphicx}
\usepackage[utf8]{inputenc}
\usepackage{lineno,hyperref}
\usepackage{multirow}
\usepackage{xcolor}
\usepackage{colortbl}
\usepackage{biblatex}
\usepackage{subcaption} 
\usepackage{adjustbox}
\definecolor{LightRed}{rgb}{1,0.75,0.75}
\definecolor{LightGreen}{rgb}{0.75,1,0.75}
\definecolor{LightBlue}{rgb}{0.75,0.75,1}

\title{\LARGE \bf
MipSLAM: Alias-Free Gaussian Splatting SLAM
}

\author{Yingzhao Li$^{1,2}$, Yan Li$^{3}$, Shixiong Tian$^{1}$, Yanjie Liu$^{1}$, Lijun Zhao$^{1,2,*}$, Gim Hee Lee$^{3}$  
\thanks{$^{*}$ Corresponding author: Lijun Zhao (zhaolj@hit.edu.cn).}%
\thanks{$^{1}$ Yingzhao Li, Shixiong Tian, Yanjie Liu and Lijun Zhao are with the State Key Laboratory of Robotics and Systems (HIT), Harbin Institute of Technology.}%
\thanks{$^{2}$ Yingzhao Li and Lijun Zhao are with Yangtze River Delta HlT Robot Technology Research Institute.}%
\thanks{$^{3}$ Yan Li and Gim Hee Lee are with Department of Computer Science, National University of Singapore.}%
\thanks{This work was supported by Heilongjiang Province's Key R\&D Program: ``Leading the Charge with Open Competition'' (Grant No. 2023ZXJ01A02) and the Key Special Projects of Heilongjiang Province's Key R\&D Program (Grant No. 2023ZX01A01).}%
}

\begin{document}

\maketitle
\thispagestyle{empty}
\pagestyle{empty}

\begin{abstract}
This paper introduces MipSLAM, a frequency-aware 3D Gaussian Splatting (3DGS) SLAM framework capable of high-fidelity anti-aliased novel view synthesis and robust pose estimation under varying camera configurations. Existing 3DGS-based SLAM systems often suffer from aliasing artifacts and trajectory drift due to inadequate filtering and purely spatial optimization. To overcome these limitations, we propose an Elliptical Adaptive Anti-aliasing (EAA) algorithm that approximates Gaussian contributions via geometry-aware numerical integration, avoiding costly analytic computation. Furthermore, we present a Spectral-Aware Pose Graph Optimization (SA-PGO) module that reformulates trajectory estimation in the frequency domain, effectively suppressing high-frequency noise and drift through graph Laplacian analysis. Extensive evaluations on Replica and TUM datasets demonstrate that MipSLAM achieves state-of-the-art rendering quality and localization accuracy across multiple resolutions.

\end{abstract}

\section{INTRODUCTION}

3D Gaussian Splatting (3DGS)~\cite{b1} has demonstrated impressive performance in high-fidelity 3D reconstruction and novel view synthesis~\cite{b2,b3}, leading to its widespread adoption in autonomous robotics and virtual/augmented reality systems. Unlike traditional pipelines~\cite{b1,b2} that require a separate structure and motion initialization stage~\cite{b4}, 3DGS-based SLAM systems~\cite{b5,b6,b24} jointly estimate camera poses and Gaussian maps within a unified architecture. However, a fundamental limitation is that the reconstructed maps are inherently tied to the fixed camera parameters used in the SLAM process. As a result, they generally produce severe artifacts when camera settings change, making alias-free Gaussian Splatting SLAM an open and critical challenge.

\begin{figure}[t]
    \captionsetup[subfigure]{font=small}
    \centering
    \begin{subfigure}{0.49\linewidth}
        \centering
        \includegraphics[width=\linewidth]{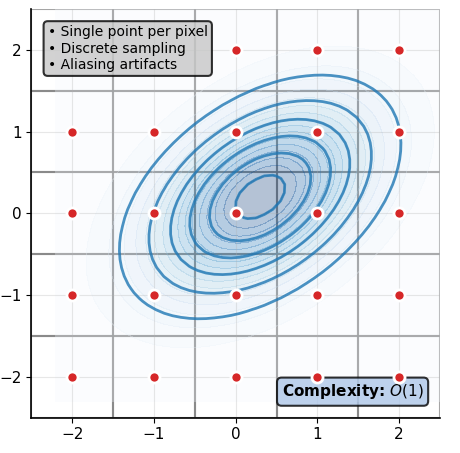} 
        \caption{Point Sampling\cite{b1} → Aliasing}
    \end{subfigure}
    \hfill
    \begin{subfigure}{0.49\linewidth}
        \centering
        \includegraphics[width=\linewidth]{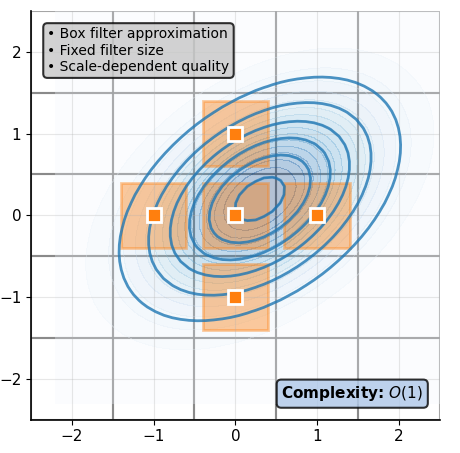}
        \caption{Box Filter\cite{b8} → Limited at high-resolution}
    \end{subfigure}
    
    \begin{subfigure}{0.49\linewidth}
        \centering
        \includegraphics[width=\linewidth]{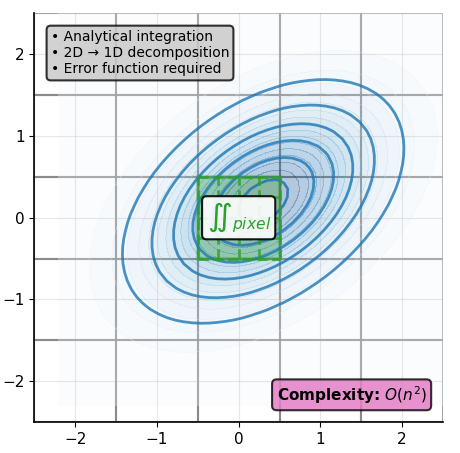}
        \caption{2D Integration\cite{b33} → High computational cost and memory}
    \end{subfigure}
    \hfill
    \begin{subfigure}{0.49\linewidth}
        \centering
        \includegraphics[width=\linewidth]{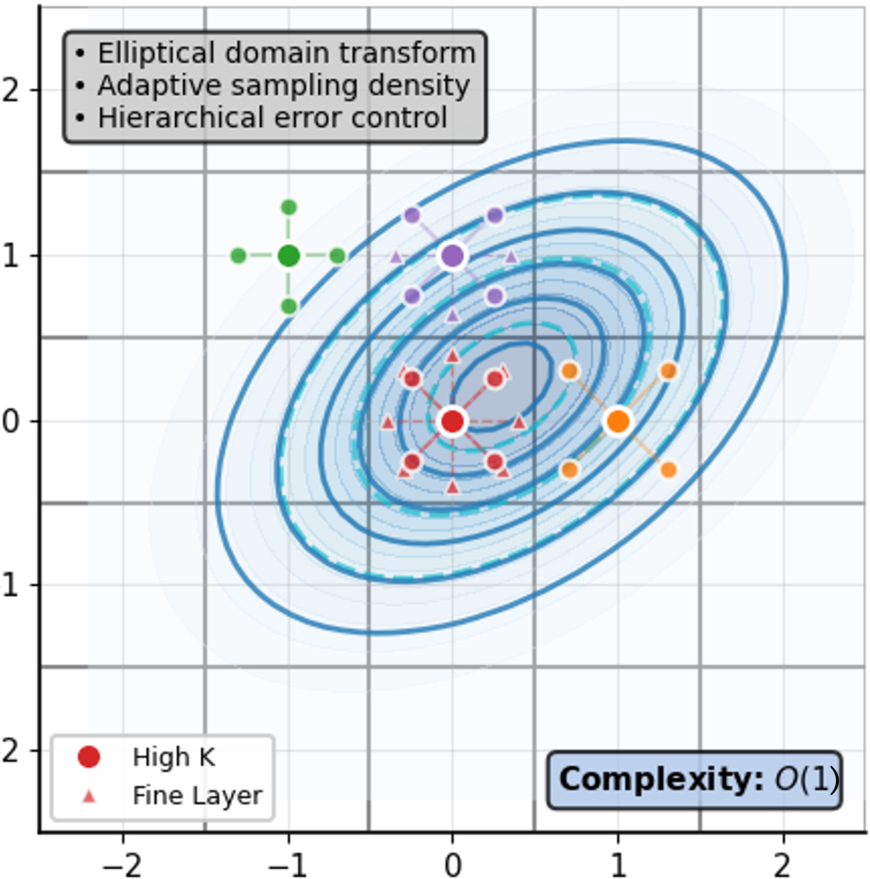} 
        \caption{Ours → Faithful and efficient rendering}
    \end{subfigure}
    
    \caption{Our method reformulates pixel rendering as a continuous integration process via stratified importance sampling, ensuring antialiasing quality and computational efficiency. The dashed ellipses depict Gaussian principal axes, color-coded by condition number: red (high), orange (moderate), purple (boundary enhancement), and green (low complexity).}
    \label{fig:qmc}
\end{figure}

RGB-D 3DGS-based SLAM frameworks such as MonoGS~\cite{b5} and SplaTAM~\cite{b6} leverage photometric and geometric constraints to track cameras and jointly optimize viewpoints and Gaussian radiance fields. In the monocular setting, Photo-SLAM~\cite{b41} emphasizes photometric consistency, while SplatSLAM~\cite{sandstrom2025splat} achieves globally consistent mapping through pose graph optimization. Although these approaches extend 3DGS to SLAM, they face challenges in camera relocalization across varying configurations, commonly suffering from rendering artifacts.

A conventional strategy to address aliasing is to retrain the Gaussian maps using alias-free splatting. Since aliasing fundamentally arises from violations of the Nyquist--Shannon sampling theorem~\cite{b7}, MipSplatting~\cite{b8} alleviates it by enforcing a minimum projected Gaussian size through 2D and 3D filtering, while Analytic-Splatting~\cite{b33} computes per-pixel contributions through analytic 2D integration.
However, for 3DGS-based SLAM systems where thousands of Gaussian primitives~\cite{b5} are continuously projected during tracking, naively incorporating these strategies~\cite{b8,b33} significantly complicates Gaussian optimization and often degrades pose estimation accuracy.

\begin{figure*}[t]
    \centering
    \includegraphics[width=0.95\linewidth]{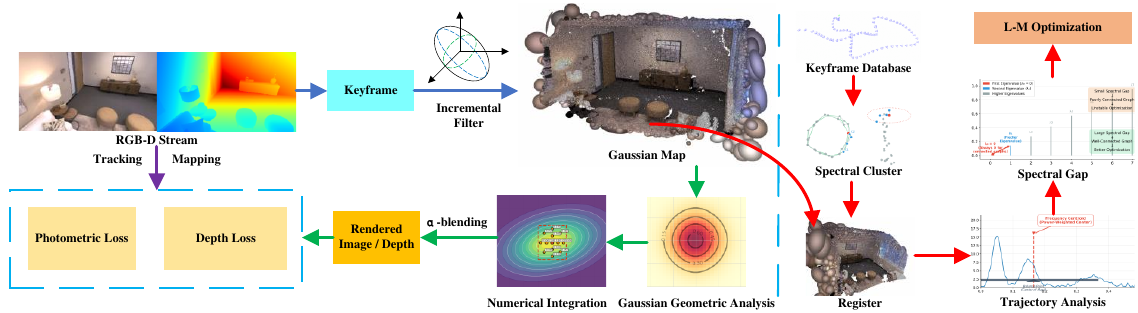}
    \caption{MipSLAM tracks RGB-D streams, estimates per-frame poses, selects keyframes, and projects Gaussian primitives into a global map. It incrementally computes 3D filter scales, compares rendered and GT RGB-D in a sliding window, and jointly optimizes poses and maps via gradient backpropagation. For $\alpha$-blending, we use elliptical projection and condition number analysis to reduce aliasing via numerical integration. SA-PGO corrects localization drift from anti-aliasing.}
    \label{fig:framework}
\end{figure*}

In this paper, we introduce MipSLAM, the first frequency-aware 3DGS SLAM framework that supports map reuse across different camera configurations (e.g., intrinsics, resolution, zoom) with high-fidelity anti-aliasing. 
First, we propose an Elliptical Adaptive Antialiasing (EAA) algorithm that accurately approximates Gaussian contributions without expensive analytic integrals. As shown in Fig.~\ref{fig:qmc}, unlike prior methods that rely on rectangular sampling, our technique projects Gaussians into an elliptical domain and adaptively selects sample points and weights, prioritizing boundary regions, based on geometric properties. This strategy efficiently approximates the integral under the Riemann framework, balancing accuracy and computational cost.
Second, we present a Spectral-Aware PGO (SA-PGO) module that reformulates trajectory optimization in the frequency domain. Rather than treating poses independently, SA-PGO models the entire trajectory as a spatiotemporal signal and applies graph Laplacian spectral decomposition to construct a frequency-consistency-driven information matrix, shifting PGO from purely spatial to joint spectral-spatial optimization. 
MipSLAM achieves superior rendering quality compared to state-of-the-art methods, while maintaining robustness across resolution changes and improving tracking accuracy.
In summary, our contributions are:
\begin{itemize}
\item We present the first frequency-aware 3D Gaussian Splatting SLAM system that supports arbitrary camera reconfiguration and eliminates aliasing artifacts. Our EAA algorithm uses adaptive numerical integration guided by Gaussian geometry, achieving high accuracy with manageable computational cost.

\item We propose SA-PGO, a spectral-analytic pose graph optimization method that models camera trajectories as spatiotemporal signals. By analyzing spectral properties and graph connectivity, SA-PGO effectively suppresses drift and improves pose estimation consistency.

\item Experiments show that MipSLAM achieves consistent high-fidelity rendering across multiple resolutions, supporting camera setups with arbitrary resolution.
\end{itemize}

\section{Related Work}

Novel View Synthesis (NVS) generates photorealistic images from unseen viewpoints. While NeRF \cite{b9} pioneered implicit scene representation via MLPs \cite{b36, b37, b38, b39}, its computational inefficiency motivated explicit alternatives. 3DGS \cite{b1} achieves real-time rendering with comparable quality through anisotropic Gaussian primitives, with extensions including GeoGaussian's geometry-aware densification \cite{b2} and Scaffold-GS's neural-anchored adaptive structures \cite{b3}. Persistent challenges involve mitigating resolution-dependent high-frequency artifacts \cite{b8}.

3DGS-based SLAM systems address the reliance on offline pose initialization (e.g., COLMAP \cite{b4}) through online geometric constraints. Key approaches include SplaTAM \cite{b6} (visibility-aware drift suppression), MonoGS \cite{b5} (multi-camera compatibility), and Gaussian-SLAM \cite{b40} (active submap optimization). These methods rely on photometric and depth losses for pose estimation and are prone to trajectory drift under challenging conditions.

Anti-aliasing techniques employ supersampling or prefiltering \cite{b16}. Mip-NeRF \cite{b17} mitigates aliasing via integrated positional encoding, while EWA splatting \cite{b18} enforces spectral constraints through anisotropic projection. Mip-Splatting \cite{b8} introduces a dual-filter architecture for Nyquist-compliant scaling and pixel-accurate projection. Analytic-Splatting \cite{b33} achieves exact pixel-window integration through eigendecomposition-based analytical integrals, but incurs quadratic complexity. Constrained by limitations in either accuracy or computational efficiency, such methods are often impeded from direct integration into SLAM systems.

\section{System Architecture}\label{sec:method}

As illustrated in Fig. \ref{fig:framework}, the architecture of MipSLAM has three main modules: tracking, alias-free Gaussian mapping, and frequency-domain pose graph optimization. The system employs the 3D Gaussian ellipsoid as its foundational representation (Sec. \ref{sec:preliminaries}) to establish a global map,
formulates an adaptive numerical integration algorithm to achieve anti-aliasing (Sec. \ref{sec:antialiasing}), 
implements SA-PGO for pose refinement (Sec. \ref{sec:sapgo}), and conducts joint optimization of both tracking and mapping components (Sec. \ref{sec:loss}).

\subsection{Preliminaries: 3D Gaussian Splatting}\label{sec:preliminaries}

We represent scenes as collections of anisotropic 3D Gaussians $\{\mathcal{G}_i\}_{i=1}^N$, where each primitive $\mathcal{G}_i$ is parameterized by position $\boldsymbol{\mu}_i \in \mathbb{R}^3$, covariance matrix $\boldsymbol{\Sigma}_i = \mathbf{O}_i \text{diag}(\mathbf{s}_i)^2 \mathbf{O}_i^T$ with rotation $\mathbf{O}_i \in SO(3)$ and scale $\mathbf{s}_i \in \mathbb{R}^3$, opacity $\alpha_i \in [0,1]$, and spherical harmonics coefficients $\mathbf{c}_i$ for view-dependent appearance. The 3D Gaussian is defined as: $ \mathcal{G}_i(\mathbf{x}) = \exp\left(-\frac{1}{2}(\mathbf{x}-\boldsymbol{\mu}_i)^T \boldsymbol{\Sigma}_i^{-1} (\mathbf{x}-\boldsymbol{\mu}_i)\right) $ where $\mathbf{x} \in \mathbb{R}^3$ denotes spatial coordinates.

Following~\cite{b1}, pixel color at coordinate $\mathbf{u} = (u,v)^T$ is computed by depth-ordered blending:
\begin{equation}
\mathbf{C}(\mathbf{u}) = \sum_{i=1}^N \mathbf{c}_i \, \alpha_i \, \mathcal{G}_i^{2D}(\mathbf{u}) \prod_{j=1}^{i-1} \bigl(1 - \alpha_j \, \mathcal{G}_j^{2D}(\mathbf{u})\bigr),
\label{eq:alpha_compositing}
\end{equation}
where $\mathbf{c}_i$ is the view-dependent color and $\alpha_i$ the learned opacity. The per-pixel depth is obtained analogously by replacing $\mathbf{c}_i$ with the camera-frame depth $z_i$ of each Gaussian:
\begin{equation}
\hat{D}(\mathbf{u}) = \sum_{i=1}^N z_i \, \alpha_i \, \mathcal{G}_i^{2D}(\mathbf{u}) \prod_{j=1}^{i-1} \bigl(1 - \alpha_j \, \mathcal{G}_j^{2D}(\mathbf{u})\bigr).
\label{eq:depth_rendering}
\end{equation}

In both expressions, the critical quantity is the evaluation of $\mathcal{G}_i^{2D}(\mathbf{u})$, which standard 3DGS~\cite{b1} computes by point-sampling the projected Gaussian at the pixel center. This point evaluation violates the Nyquist--Shannon sampling theorem~\cite{b7} when the camera resolution changes, producing aliasing artifacts. We replace this point evaluation with a principled numerical integration over the pixel footprint: the per-pixel contribution $\alpha_i\mathcal{G}_i^{2D}(\mathbf{u})$ in Eqs.~\eqref{eq:alpha_compositing}--\eqref{eq:depth_rendering} is replaced by the integrated opacity $\alpha$ computed in Eq.~\eqref{eq:numerical_integration_compact}.

\subsection{Elliptical Adaptive Anti-aliasing}\label{sec:antialiasing}

We propose a {geometry-driven elliptical sampling strategy} that approximates {analytical integration accuracy} through {importance-weighted numerical quadrature}, achieving superior antialiasing performance with provable convergence guarantees. A comparative analysis against alternative approaches is depicted in Fig. \ref{fig:qmc}.

\subsubsection{Incremental 3D Filter}

To build a Gaussian map amenable to increased resolution, we apply 3D filtering \cite{b8} within a sliding window of keyframes, avoiding the prohibitive cost of processing all frames. Let $\mathcal{K}_t$ be the active keyframes at time $t$ and $\mathcal{V}(\mathcal{G}_k)$ the visibility set of Gaussian $\mathcal{G}_k$. We maintain a covisibility graph $\mathcal{G}_c = (\mathcal{K}, \mathcal{E})$ where edges $\mathcal{E}$ connect mutually visible keyframes. The sampling frequency $\hat{\nu}_k^{t}$ for Gaussian $k$ is updated as:
\begin{equation}
    \hat{\nu}_k^{t+1} = \max\left( \hat{\nu}_k^t, \max_{n \in \mathcal{V}(\mathcal{G}_k)} \left\{ \frac{f_n}{d_n} \right\} \right),
    \label{eq:freq_update}
\end{equation}
where $f_n$ is the focal length and $d_n$ is the depth of Gaussian $k$ observed from keyframe $n$.

\subsubsection{Geometry-driven Elliptical-domain Sampling} 

Employing a simple box filter \cite{b8} is insufficient to address resolution reduction or ``zooming out'', as rectangular pixel sampling misaligns with the natural elliptical geometry of projected 2D Gaussians, introducing systematic bias in alpha computation. We reformulate the pixel integral by transforming from pixel coordinates to the Gaussian's intrinsic elliptical domain. For a Gaussian with 2D projected covariance $\boldsymbol{\Sigma}^{2D}$ and projected mean $\boldsymbol{\mu}^{2D}$, the integrated intensity over a pixel footprint $P$ is~\cite{b33}:
\begin{equation}
I(P) = \iint_P \alpha_i \exp\!\left(-\frac{1}{2} \Delta\mathbf{x}^T (\boldsymbol{\Sigma}^{2D})^{-1} \Delta\mathbf{x} \right) d\mathbf{x},
\label{eq:pixel_integration}
\end{equation}
where $\Delta\mathbf{x} = \mathbf{x} - \boldsymbol{\mu}^{2D}$ is the displacement from the Gaussian center and $\alpha_i$ is the learned opacity. Via eigenvalue decomposition $\boldsymbol{\Sigma}^{2D} = \mathbf{Q}\boldsymbol{\Lambda}\mathbf{Q}^T$ with eigenvalue matrix $\boldsymbol{\Lambda} = \text{diag}(\lambda_1, \lambda_2)$ and orthonormal eigenvector matrix $\mathbf{Q} = [\mathbf{e}_1, \mathbf{e}_2]$, we define the principal-axis coordinates $\mathbf{v}_k = \mathbf{Q}^T(\mathbf{x}_k - \boldsymbol{\mu}^{2D})$ for each sample point $\mathbf{x}_k$. This transformation aligns the integration domain with the Gaussian's principal axes, eliminating sampling bias and enabling optimal quadrature along iso-probability contours.

Highly anisotropic or boundary-proximate Gaussians exhibit large condition numbers $\kappa = \sqrt{\lambda_{\max}/\lambda_{\min}}$, demanding adaptive sampling density. {We increase sampling rate with $\kappa$ } to capture fine geometric details while maintaining analytical integration fidelity \cite{b33}. Boundary pixels are detected via a distance field, with sample weighting emphasizing regions near discontinuities.

To achieve optimal convergence to the analytical integral, {we employ importance sampling that concentrates quadrature} points in regions of maximum contribution variance. For each sample point $\mathbf{x}_k$ with principal-axis coordinates $\mathbf{v}_k$, the importance weight is computed as:

\begin{equation}
w_k = \frac{q(\mathbf{x}_k)}{p(\mathbf{x}_k)} \cdot \psi(\kappa, \mathbf{x}_k),
\label{eq:importance_weights}
\end{equation}
where $q(\mathbf{x}_k) = \exp(-\frac{1}{2}\|\boldsymbol{\Lambda}^{-1/2}\mathbf{v}_k\|^2)$ is the target sampling density proportional to the Gaussian evaluated at the sample, $p(\mathbf{x}_k)$ is the proposal density from which samples are drawn, and $\psi(\kappa, \mathbf{x}_k) = (1 + \gamma\log\kappa)\phi_b(\mathbf{x}_k)$ is the geometry-boundary enhancement factor. Here $\gamma$ is a scaling coefficient, and $\phi_b(\mathbf{x}_k) = \exp(-\beta d_b(\mathbf{x}_k))$ encodes proximity to pixel boundaries via the Euclidean distance $d_b(\mathbf{x}_k)$ from $\mathbf{x}_k$ to the nearest pixel edge, with decay rate $\beta$.

\begin{figure}[t]
    \captionsetup[subfigure]{font=small}
    \centering
    \begin{subfigure}{0.49\linewidth}
        \centering
        \includegraphics[width=\linewidth]{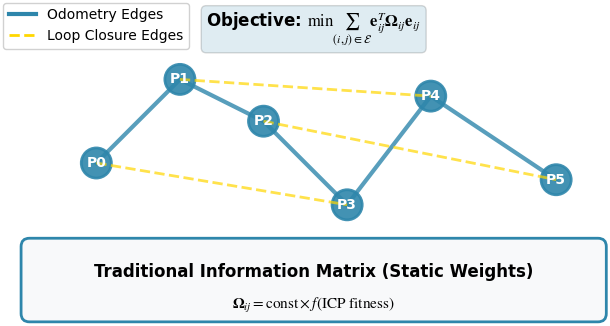} 
        \caption{Traditional PGO}
    \end{subfigure}
    \hfill
    \begin{subfigure}{0.49\linewidth}
        \centering
        \includegraphics[width=\linewidth]{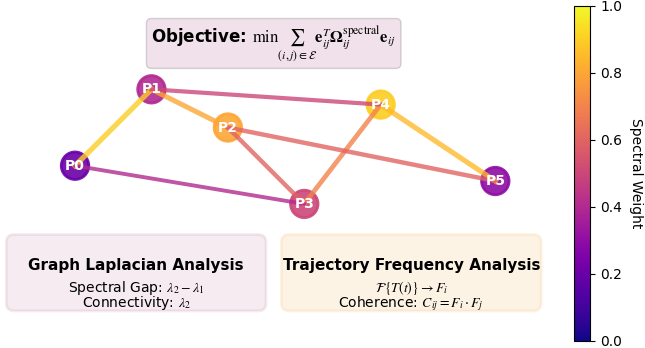}
        \caption{Spectral-Aware PGO}
    \end{subfigure}
    \hfill
    
    \begin{subfigure}{\linewidth}
        \centering
        \includegraphics[width=\linewidth]{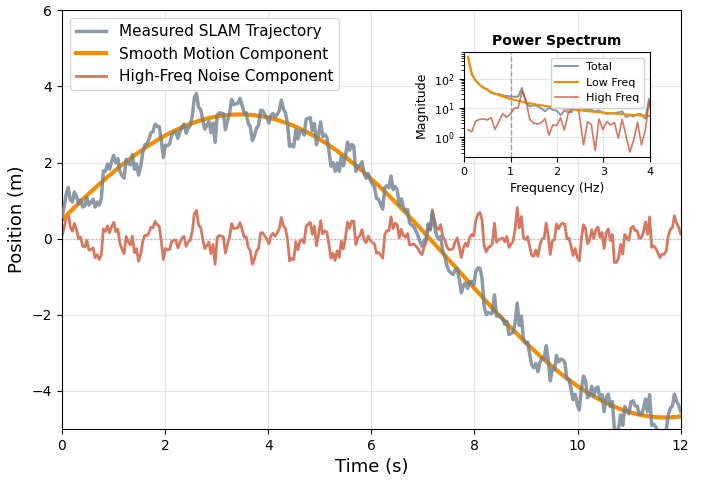}
        \caption{Trajectory FFT Analysis}
    \end{subfigure}
    \hfill
    \begin{subfigure}{\linewidth}
        \centering
        \includegraphics[width=\linewidth]{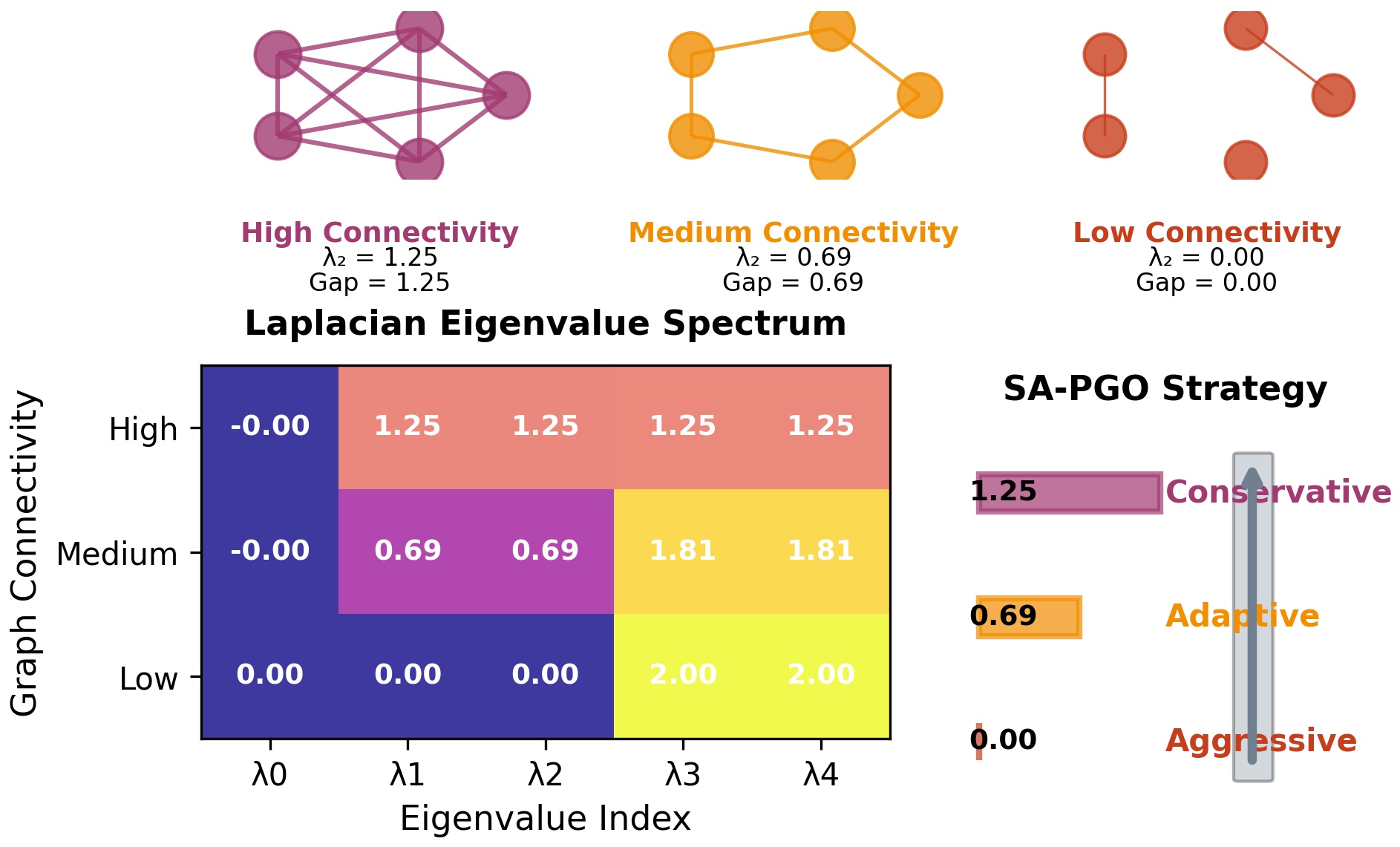} 
        \caption{Laplacian Decomposition}
    \end{subfigure}
    
    \caption{SA-PGO leverages frequency-domain trajectory analysis and graph spectral theory to achieve adaptive noise suppression and intelligent spectral-aware optimization.}
    \label{fig:pgo}
\end{figure}

\subsubsection{Numerical Integration Approximation}

The numerical integration employs Gaussian quadrature principles \cite{b34} adapted for elliptical domains. Using the principal-axis coordinates $\mathbf{v}_k = \mathbf{Q}^T(\mathbf{x}_k - \boldsymbol{\mu}^{2D})$, the integrated opacity $\alpha$ that replaces the point-sampled $\alpha_i\mathcal{G}_i^{2D}(\mathbf{u})$ in Eqs.~\eqref{eq:alpha_compositing}--\eqref{eq:depth_rendering} is approximated as:

\begin{equation}
\alpha = \frac{\alpha_i}{\sum_k w_k} \sum_{k=1}^K w_k \exp\!\left( -\frac{1}{2} \mathbf{v}_k^\top \boldsymbol{\Lambda}^{-1} \mathbf{v}_k \right),
\label{eq:numerical_integration_compact}
\end{equation}
where $K$ is the number of quadrature points, $\boldsymbol{\Lambda} = \text{diag}(\lambda_1, \lambda_2)$ is the eigenvalue matrix of $\boldsymbol{\Sigma}^{2D}$, and $w_k$ are the importance weights from Eq.~\eqref{eq:importance_weights}.

{Our method extends traditional Gaussian--Legendre quadrature \cite{b35} to anisotropic elliptical domains,} combining provable accuracy guarantees with computational efficiency through stratified importance sampling that adapts to both geometric anisotropy and boundary proximity.

\subsubsection{Gradient Computation}

Gradient computation maintains exact correspondence with the importance-weighted numerical integration through automatic differentiation. Let $\boldsymbol{\Theta}$ denote the differentiable parameter set (including $\boldsymbol{\mu}^{2D}$ and $\boldsymbol{\Sigma}^{2D}$), and define $P_k = -\frac{1}{2}\mathbf{v}_k^\top \boldsymbol{\Lambda}^{-1} \mathbf{v}_k$ as the quadratic form at sample point $k$. The gradient of the integrated opacity with respect to $\boldsymbol{\Theta}$ is:
\begin{equation}
\frac{\partial \alpha}{\partial \boldsymbol{\Theta}} = \frac{\alpha_i}{\sum_{k} w_k} \sum_{k=1}^{K} w_k \exp(P_k) \frac{\partial P_k}{\partial \boldsymbol{\Theta}}.
\label{eq:cess_gradients}
\end{equation}
This ensures stable backpropagation with consistent convergence properties between forward and backward passes.

\subsection{Spectral-Aware Pose Graph Optimization}\label{sec:sapgo}

As shown in Fig. \ref{fig:pgo}, we propose SA-PGO, which integrates frequency-domain trajectory analysis with graph spectral theory to achieve adaptive optimization. The module operates on the pose graph produced by the SLAM front-end and outputs refined pose estimates $\mathbf{x}^*$ (Eq.~\eqref{eq:sapgo_objective}), which are then used to update the camera extrinsics for joint Gaussian map optimization (Sec.~\ref{sec:loss}).

\subsubsection{Multi-Modal Descriptor Extraction}

For input image $I(x,y)$, we extract frequency features $\mathbf{f}_{freq}$ through the 2D FFT \cite{b29}, gradient features $\mathbf{f}_{grad}$ using multi-scale Sobel operators, texture features $\mathbf{f}_{tex}$ via multiple convolution kernels, and color features $\mathbf{f}_{color}$ from channel-wise histograms:
\begin{equation}
\mathbf{d} = [\mathbf{f}_{freq}^T, \mathbf{f}_{grad}^T, \mathbf{f}_{tex}^T, \mathbf{f}_{color}^T]^T \in \mathbb{R}^{d_{total}},
\label{eq:multimodal_descriptor}
\end{equation}
where $d_{total}$ is the total descriptor dimension, and each component is L2-normalized within its respective subspace to maintain balanced contribution across modalities.

\subsubsection{Frequency-domain Trajectory Analysis}
 
A well-behaved SLAM trajectory should exhibit dominant low-frequency components with suppressed high-frequency oscillations. For pose sequence $\{\mathbf{p}_1, \ldots, \mathbf{p}_N\}$ where $\mathbf{p}_i = [\mathbf{t}_i^T, \boldsymbol{\omega}_i^T]^T \in \mathbb{R}^6$ represents translation and rotation (axis-angle), we compute the sliding-window DFT \cite{b29}:
\begin{equation}
\mathcal{F}_k^{(d)}(\omega) = \sum_{n=1}^{N_w} p_{k+n-1}^{(d)} \, w(n) \, e^{-j2\pi\omega(n-1)/N_w},
\label{eq:trajectory_dft}
\end{equation}
where $k$ is the starting frame index, $N_w$ is the sliding-window size, $d \in \{1,\ldots,6\}$ indexes the six pose components, $p_{k+n-1}^{(d)}$ is the $d$-th component of pose $\mathbf{p}_{k+n-1}$, $\omega$ is the discrete frequency index, and $w(n)$ is a Hanning window applied to reduce spectral leakage. The frequency signature \cite{b30} captures the dominant spectral characteristics through the weighted frequency centroid:
\begin{equation}
\mathbf{S}_k = \left[\frac{\sum_{\omega=0}^{\omega_{max}} \omega \cdot |\mathcal{F}_k^{(d)}(\omega)|^2}{\sum_{\omega=0}^{\omega_{max}} |\mathcal{F}_k^{(d)}(\omega)|^2 + \epsilon_{reg}}\right]_{d=1}^{6},
\label{eq:spectral_signature}
\end{equation}
where $\omega_{max}$ is the maximum frequency index, $|\mathcal{F}_k^{(d)}(\omega)|^2$ is the power spectral density, and $\epsilon_{reg}$ prevents numerical instability. A higher centroid value indicates more high-frequency energy, signaling potential drift or noise.

The frequency coherence between poses $i$ and $j$ quantifies trajectory smoothness consistency:
\begin{equation}
\mathcal{C}_{ij} = \frac{1}{2}\left(1 + \frac{\mathbf{S}_i \cdot \mathbf{S}_j}{\|\mathbf{S}_i\|_2 \|\mathbf{S}_j\|_2 + \epsilon_{reg}}\right).
\label{eq:freq_coherence}
\end{equation}

For the relative transformation $\mathbf{T}_{ij}$ between poses $i$ and $j$ with translation $\mathbf{t}_{ij}$ and rotation $\boldsymbol{\omega}_{ij}$, we compute geometric regularity $\mathcal{G}_{ij} = \exp(-\alpha_t \|\mathbf{t}_{ij}\|_2 - \alpha_r \|\boldsymbol{\omega}_{ij}\|_2)$. The overall spectral confidence combines frequency coherence and geometric regularity:
\begin{equation}
\mathcal{S}_{ij} = \beta_c \, \mathcal{C}_{ij} + \beta_g \, \mathcal{G}_{ij},
\label{eq:spectral_confidence}
\end{equation}
where $\beta_c + \beta_g = 1$ are balancing weights.

\subsubsection{Graph Laplacian Spectral Decomposition}

SA-PGO leverages graph spectral theory \cite{b31} to assess pose graph connectivity and guide optimization strategies. We construct the weighted adjacency matrix $\mathbf{A} \in \mathbb{R}^{n \times n}$ where $A_{ij} = \mathcal{S}_{ij}$ for connected poses $i$ and $j$ (and zero otherwise), with $n$ being the number of poses. The degree matrix $\mathbf{D}$ is diagonal with $D_{ii} = \sum_j A_{ij}$. The normalized graph Laplacian is:
\begin{equation}
\mathbf{L}_{norm} = \mathbf{D}^{-1/2}(\mathbf{D} - \mathbf{A})\mathbf{D}^{-1/2}.
\label{eq:laplacian}
\end{equation}

The eigendecomposition $\mathbf{L}_{norm} = \mathbf{V}\boldsymbol{\Lambda}_L\mathbf{V}^T$ yields eigenvalues $0 = \lambda_1^L \leq \lambda_2^L \leq \cdots \leq \lambda_n^L$ and corresponding eigenvectors $\mathbf{V} = [\mathbf{v}_1, \ldots, \mathbf{v}_n]$. The spectral gap $\delta = \lambda_2^L$ (the Fiedler eigenvalue~\cite{b31}) measures algebraic connectivity: a larger $\lambda_2^L$ indicates better graph connectivity and more stable optimization. When $\lambda_2^L < \tau_{opt}$ (poor connectivity), we increase edge pruning thresholds to maintain stability; when $\lambda_2^L > \tau_{opt}$, we enable aggressive spectral-guided optimization with enhanced edge reweighting.

\subsubsection{Spectral Edge Selection and Regularized PGO}

SA-PGO incorporates non-odometry constraints through spectral candidate selection. Loop-closure and adjacent edges are identified using spectral clustering \cite{b32} on the first $k$ non-trivial Laplacian eigenvectors $\mathbf{V}_k = [\mathbf{v}_2, \ldots, \mathbf{v}_{k+1}]$, where $k$ is the number of clusters. Candidate pose pairs $(i,j)$ are filtered by requiring spectral coherence $\mathcal{C}_{ij} \geq \tau_{freq}$ and evaluating a hybrid similarity $S_{ij} = S_{cos}(\mathbf{d}_i, \mathbf{d}_j) + S_{euc}(\mathbf{d}_i, \mathbf{d}_j)$ between multi-modal descriptors from Eq.~\eqref{eq:multimodal_descriptor}.

The full optimization objective combines a geometric term with two spectral regularizers:
\begin{equation}
\mathbf{x}^* = \arg\min_{\mathbf{x}}\bigl(E_{geo}(\mathbf{x}) + \lambda_{spe}\, R_{spe}(\mathbf{x}) + \lambda_{smo}\, R_{smo}(\mathbf{x})\bigr),
\label{eq:sapgo_objective}
\end{equation}
where $\mathbf{x} = \{\mathbf{T}_1, \ldots, \mathbf{T}_n\}$ collects all $SE(3)$ camera poses, $E_{geo}(\mathbf{x}) = \sum_{(i,j) \in \mathcal{E}} \| \log(\mathbf{T}_{ij}^{-1} \mathbf{T}_i^{-1} \mathbf{T}_j) \|_{\boldsymbol{\Omega}_{ij}}^2$ is the standard pose graph residual with information matrices $\boldsymbol{\Omega}_{ij}$ weighted by spectral confidence $\mathcal{S}_{ij}$, $R_{spe}(\mathbf{x})$ penalizes inconsistency between spectrally similar poses, $R_{smo}(\mathbf{x})$ enforces temporal smoothness by penalizing high-frequency oscillations, and $\lambda_{spe}, \lambda_{smo} \geq 0$ are regularization weights. We adapt $\lambda_{spe}$ and $\lambda_{smo}$ using the spectral gap: a higher $\lambda_2^L$ increases spectral influence, while a lower value prioritizes geometric fidelity.

\subsection{Loss Function}\label{sec:loss}

The system jointly optimizes camera poses $\mathbf{T} \in SE(3)$ and Gaussian parameters $\boldsymbol{\Theta}_G = \{\boldsymbol{\mu}_i, \boldsymbol{\Sigma}_i, \alpha_i, \mathbf{c}_i\}$ through the combined loss:
\begin{equation}
\mathcal{L} = \lambda_{rgb}\mathcal{L}_{rgb} + \lambda_{depth}\mathcal{L}_{depth},
\label{eq:jointloss}
\end{equation}
where $\mathcal{L}_{rgb} = \|\mathbf{C}(\mathbf{u}) - \mathbf{I}_{gt}(\mathbf{u})\|_2^2$ and $\mathcal{L}_{depth} = \|\hat{D}(\mathbf{u}) - D_{gt}(\mathbf{u})\|_2^2$ are photometric and depth losses, respectively. This per-frame optimization runs asynchronously with SA-PGO (Eq.~\eqref{eq:sapgo_objective}), which periodically refines the global pose graph in the back-end.

\begin{table*}[t]
\centering
\small
\setlength{\tabcolsep}{4pt} 
\begin{tabularx}{\linewidth}{ 
  >{\centering\arraybackslash}p{0.2cm} 
  >{\centering\arraybackslash}p{2.1cm}
  @{\hspace{2pt}}
  *{18}{>{\centering\arraybackslash}X} 
  }
\toprule
\multicolumn{2}{c}{\multirow{2}{*}{Method}} & 
\multicolumn{6}{c}{PSNR $\uparrow$} & 
\multicolumn{6}{c}{SSIM $\uparrow$} & 
\multicolumn{6}{c}{LPIPS $\downarrow$} \\ 
\cmidrule(lr){3-8} \cmidrule(lr){9-14} \cmidrule(lr){15-20}
& & 2 & 1 & $1/2$ & $1/4$ & $1/8$ & Avg
& 2 & 1 & $1/2$ & $1/4$ & $1/8$ & Avg
& 2 & 1 & $1/2$ & $1/4$ & $1/8$ & Avg \\ 
\midrule
\multirow{2}{*}{\rotatebox[origin=c]{90}{\scriptsize Recon}} 
& 3D GS \cite{b1}
& \cellcolor{LightBlue}{38.71} & \cellcolor{LightRed}{42.20} & \cellcolor{LightBlue}{38.22} & \cellcolor{LightGreen}{29.84} & \cellcolor{LightBlue}{25.99} & \cellcolor{LightBlue}{34.99}
& \cellcolor{LightBlue}{0.974} & \cellcolor{LightRed}{0.981} & \cellcolor{LightGreen}{0.963} & 0.888 & \cellcolor{LightGreen}{0.835} & \cellcolor{LightGreen}{0.928}
& 0.108 & 0.083 & 0.069 & 0.091 & 0.121 & 0.094 \\ 

& Sca-GS \cite{b3}
& \cellcolor{LightGreen}{38.42} & \cellcolor{LightBlue}{40.97} & \cellcolor{LightGreen}{35.86} & \cellcolor{LightBlue}{29.91} & 24.28 & \cellcolor{LightGreen}{33.89}
& \cellcolor{LightRed}{0.975} & 0.968 & 0.954 & 0.884 & 0.796 & 0.915
& 0.101 & 0.085 & 0.067 & \cellcolor{LightGreen}{0.079} & 0.137 & 0.094 \\ 

\midrule

\multirow{4}{*}{\rotatebox[origin=c]{90}{\scriptsize SLAM}} 
& SplaTAM \cite{b6}
& 33.68 & 35.54 & 32.80 & 25.22 & 22.64 & 29.98
& \cellcolor{LightBlue}{0.974} & \cellcolor{LightBlue}{0.976} & \cellcolor{LightBlue}{0.966} & \cellcolor{LightBlue}{0.912} & \cellcolor{LightBlue}{0.842} & \cellcolor{LightBlue}{0.934}
& 0.124 & 0.073 & 0.064 & 0.127 & 0.123 & 0.102 \\ 

& MonoGS \cite{b5}
& 36.98 & \cellcolor{LightGreen}{39.51} & 34.67 & 28.43 & 24.01 & 32.72
& \cellcolor{LightGreen}0.969 & \cellcolor{LightGreen}{0.975} & \cellcolor{LightGreen}{0.963} & \cellcolor{LightGreen}{0.899} & 0.812 & 0.924
& \cellcolor{LightGreen}{0.077} & \cellcolor{LightRed}{0.044} & \cellcolor{LightBlue}{0.031} & \cellcolor{LightBlue}{0.071} & \cellcolor{LightBlue}{0.098} & \cellcolor{LightBlue}{0.064} \\ 

& GS-ICP \cite{b24}
& 34.00 & 37.55 & 33.64 & 28.68 & \cellcolor{LightGreen}{24.67} & 31.71
& 0.952 & 0.967 & 0.950 & 0.870 & 0.787 & 0.905
& \cellcolor{LightBlue}{0.074} & \cellcolor{LightGreen}{0.056} & \cellcolor{LightGreen}{0.044} & 0.082 & \cellcolor{LightGreen}{0.104} & \cellcolor{LightGreen}{0.072} \\ 

& Ours
& \cellcolor{LightRed}{38.73} & 39.05 & \cellcolor{LightRed}{38.56} & \cellcolor{LightRed}{33.33} & \cellcolor{LightRed}{29.47} & \cellcolor{LightRed}{35.83}
& \cellcolor{LightRed}{0.975} & 0.972 & \cellcolor{LightRed}{0.973} & \cellcolor{LightRed}{0.954} & \cellcolor{LightRed}{0.919} & \cellcolor{LightRed}{0.959}
& \cellcolor{LightRed}{0.068} & \cellcolor{LightBlue}{0.046} & \cellcolor{LightRed}{0.018} & \cellcolor{LightRed}{0.047} & \cellcolor{LightRed}{0.061} & \cellcolor{LightRed}{0.048} \\ 
\bottomrule
\end{tabularx}
\caption{Multiresolution testing results on the Replica dataset \cite{b25}. The best results are shown in \cellcolor{LightRed}{red}, the second-best results in \cellcolor{LightBlue}{blue}, and the third-best results in \cellcolor{LightGreen}{green}. The data are averages over 8 sequences. A dash (`--') indicates a system crash during tracking. }
\label{table:replica}
\end{table*}

\begin{table*}[h]
\centering
\small
\setlength{\tabcolsep}{4pt} 
\begin{tabularx}{\linewidth}{ 
  >{\centering\arraybackslash}p{2.5cm} 
  @{\hspace{2pt}}
  *{18}{>{\centering\arraybackslash}X} 
  }
\toprule
Method & 
\multicolumn{6}{c}{PSNR $\uparrow$} & 
\multicolumn{6}{c}{SSIM $\uparrow$} & 
\multicolumn{6}{c}{LPIPS $\downarrow$} \\ 
\cmidrule(lr){2-7} \cmidrule(lr){8-13} \cmidrule(lr){14-19}
& 4 & 2 & 1 & $1/2$ & $1/4$ & Avg
& 4 & 2 & 1 & $1/2$ & $1/4$ & Avg
& 4 & 2 & 1 & $1/2$ & $1/4$ & Avg \\ 
\midrule

SplaTAM \cite{b6}
& \cellcolor{LightGreen}{21.70} & \cellcolor{LightGreen}{21.78} & \cellcolor{LightGreen}{21.93} & \cellcolor{LightBlue}{21.46} & \cellcolor{LightBlue}{20.68} & \cellcolor{LightGreen}{21.51}
& 0.821 & \cellcolor{LightGreen}{0.822} & \cellcolor{LightRed}{0.785} & \cellcolor{LightGreen}{0.765} & \cellcolor{LightGreen}{0.682} & \cellcolor{LightGreen}{0.775}
& \cellcolor{LightGreen}{0.302} & \cellcolor{LightGreen}{0.289} & \cellcolor{LightRed}{0.244} & \cellcolor{LightGreen}{0.222} & 0.258 & \cellcolor{LightGreen}{0.263} \\ 

MonoGS \cite{b5}
& \cellcolor{LightBlue}{23.04} & \cellcolor{LightBlue}{23.33} & \cellcolor{LightRed}{23.57} & \cellcolor{LightGreen}{21.44} & \cellcolor{LightGreen}{20.45} & \cellcolor{LightBlue}{22.37}
& \cellcolor{LightBlue}{0.862} & \cellcolor{LightBlue}{0.823} & \cellcolor{LightBlue}{0.784} & \cellcolor{LightBlue}{0.783} & \cellcolor{LightBlue}{0.742} & \cellcolor{LightBlue}{0.799}
& \cellcolor{LightBlue}{0.275} & \cellcolor{LightBlue}{0.273} & \cellcolor{LightBlue}{0.247} & \cellcolor{LightBlue}{0.193} & \cellcolor{LightBlue}{0.147} & \cellcolor{LightBlue}{0.227} \\ 

GS-ICP  \cite{b24}
& 17.67 & 17.68 & 17.69 & 17.57 & 17.27 & 17.58
& \cellcolor{LightGreen}{0.851} & 0.780 & 0.706 & 0.659 & 0.646 & 0.728
& 0.307 & 0.317 & 0.302 & 0.282 & \cellcolor{LightGreen}{0.234} & 0.288 \\ 

Ours
& \cellcolor{LightRed}{23.34} & \cellcolor{LightRed}{23.52} & \cellcolor{LightBlue}{23.21} & \cellcolor{LightRed}{21.76} & \cellcolor{LightRed}{22.03} & \cellcolor{LightRed}{22.77}
& \cellcolor{LightRed}{0.865} & \cellcolor{LightRed}{0.827} & \cellcolor{LightGreen}{0.782} & \cellcolor{LightRed}{0.789} & \cellcolor{LightRed}{0.788} & \cellcolor{LightRed}{0.810}
& \cellcolor{LightRed}{0.271} & \cellcolor{LightRed}{0.268} & \cellcolor{LightGreen}{0.249} & \cellcolor{LightRed}{0.189} & \cellcolor{LightRed}{0.123} & \cellcolor{LightRed}{0.220} \\ 
\bottomrule
\end{tabularx}
\caption{Multiresolution testing results on the TUM dataset \cite{b26}. The best results are shown in \cellcolor{LightRed}{red}, the second-best results in \cellcolor{LightBlue}{blue}, and the third-best results in \cellcolor{LightGreen}{green}.}
\label{table:tum}
\end{table*}

\begin{table}
\centering
\resizebox{.99\linewidth}{!}{%
\begin{tabular}{@{}ll|ccccccccc@{}}
\toprule
\multicolumn{2}{c|}{Method} & R0 & R1 & R2 & Off0 & Off1 & Off2 & Off3 & Off4 & Avg \\ \midrule

\multirow{3}{*}{\rotatebox[origin=c]{90}{\scriptsize Implicit}} 
& NICE-SLAM \cite{b36} & 0.97 & 1.31 & 1.07 & 0.88 & 1.00 & 1.06 & 1.10 & 1.13 & 1.07 \\
& Point-SLAM \cite{b37} & 0.61 & 0.41 & 0.37 & \cellcolor{LightBlue}{0.38} & 0.48 & 0.54 & 0.69 & 0.72 & 0.53 \\
& GO-SLAM \cite{b38} & \cellcolor{LightGreen}{0.34} & \cellcolor{LightRed}{0.29} & \cellcolor{LightGreen}{0.30} & \cellcolor{LightRed}{0.32} & \cellcolor{LightGreen}{0.30} & 0.39 & 0.39 & \cellcolor{LightBlue}{0.35} & \cellcolor{LightBlue}{0.35} \\  

\midrule

\multirow{4}{*}{\rotatebox[origin=c]{90}{\scriptsize 3D GS}} 
& MonoGS~\cite{b5} & 0.48 & \cellcolor{LightGreen}{0.32} & 0.33 & 0.51 & 0.57 & \cellcolor{LightBlue}{0.22} & \cellcolor{LightRed}{0.18} & 2.06 & 0.58 \\  
& SplaTAM~\cite{b6} & \cellcolor{LightBlue}{0.31} & 0.40 & \cellcolor{LightBlue}{0.29} & 0.47 & \cellcolor{LightBlue}{0.27} & \cellcolor{LightGreen}{0.29} & \cellcolor{LightGreen}{0.32} & 0.72 & \cellcolor{LightGreen}{0.38} \\ 
& Photo-SLAM \cite{b41} & 0.54 & 0.39 & 0.31 & 0.52 & 0.44 & 1.28 & 0.78 & \cellcolor{LightGreen}{0.58} & 0.61 \\   
& Ours & \cellcolor{LightRed}{0.27} & \cellcolor{LightBlue}{0.30} & \cellcolor{LightRed}{0.28} & \cellcolor{LightGreen}{0.39} & \cellcolor{LightRed}{0.25} & \cellcolor{LightRed}{0.21} & \cellcolor{LightRed}{0.18} & \cellcolor{LightRed}{0.33} & \cellcolor{LightRed}{0.28} \\ 
\bottomrule
\end{tabular}}
\caption{Comparison of localization accuracy (ATE RMSE [cm] $\downarrow$) on the Replica dataset \cite{b25}.}
\label{tab:tracking}
\end{table}

\section{Experiments}

\subsection{Implementation Details}

\begin{figure}[t]
    \centering
    \includegraphics[width=\linewidth]{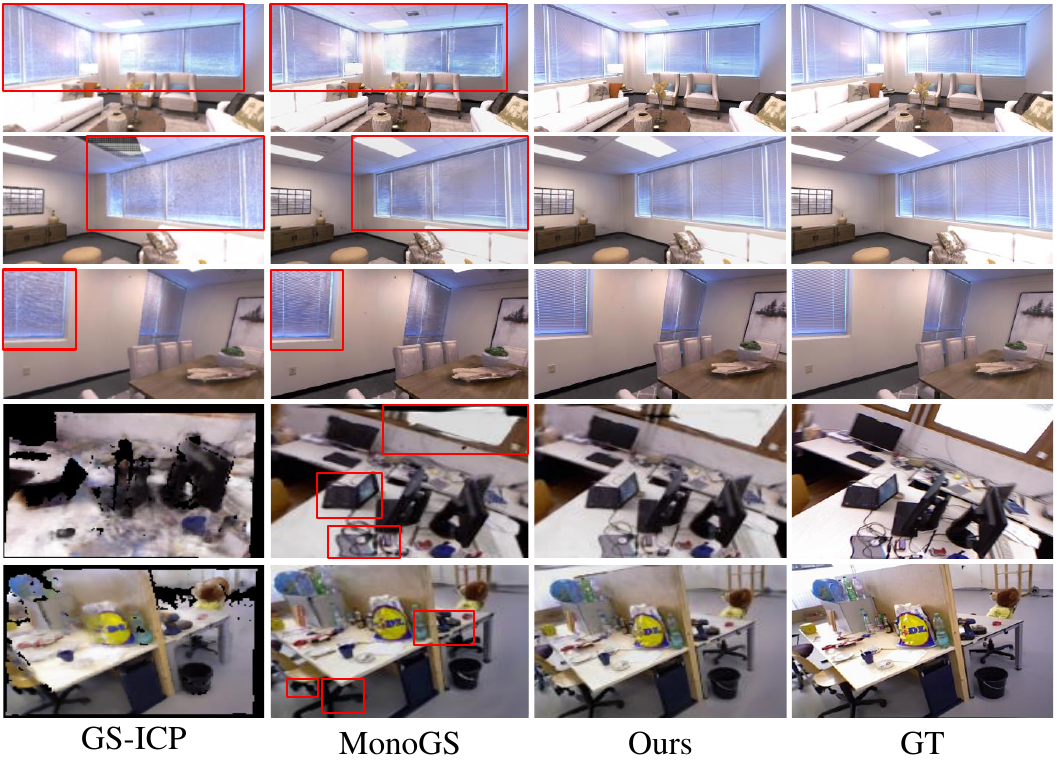}
    \caption{Testing results at 1$/$4 resolution on the Replica \cite{b25} and TUM datasets \cite{b26}. Previous methods have exhibited varying degrees of aliasing, blurring, and inflation issues. Our MipSLAM can faithfully represent the real world.}
    \label{f4}
\end{figure}

To evaluate the proposed method, MipSLAM was assessed on the Replica~\cite{b25} (8 sequences) and TUM RGB-D~\cite{b26} (3 sequences) datasets. We compared against GS-based SLAM systems (MonoGS~\cite{b5}, SplaTAM~\cite{b6}, GS-ICP~\cite{b24}) and 3DGS approaches (Vanilla 3DGS~\cite{b1}, Scaffold-GS~\cite{b3}). The localization experiment also compares some neural implicit schemes \cite{b36, b37, b38, b41}. Models were trained at native resolution and evaluated across multiple resolutions to emulate changes in camera parameters. Metrics include PSNR, SSIM, LPIPS, FPS, and ATE RMSE. All experiments were conducted on an NVIDIA RTX 4090 GPU.

Image resolutions were scaled to simulate focal length variations~\cite{b1}: downsampling introduces aliasing, while upsampling produces high-frequency artifacts. Replica (native 680$\times$1200) was evaluated at scales of 2$\times$, 1$\times$, 1/2, 1/4, and 1/8. TUM (native 480$\times$640) was evaluated at 4$\times$, 2$\times$, 1$\times$, 1/2, and 1/4.

\subsection{Multi-resolution Rendering}\label{4.2}

\begin{figure}[t]
    \centering
    \includegraphics[width=\linewidth]{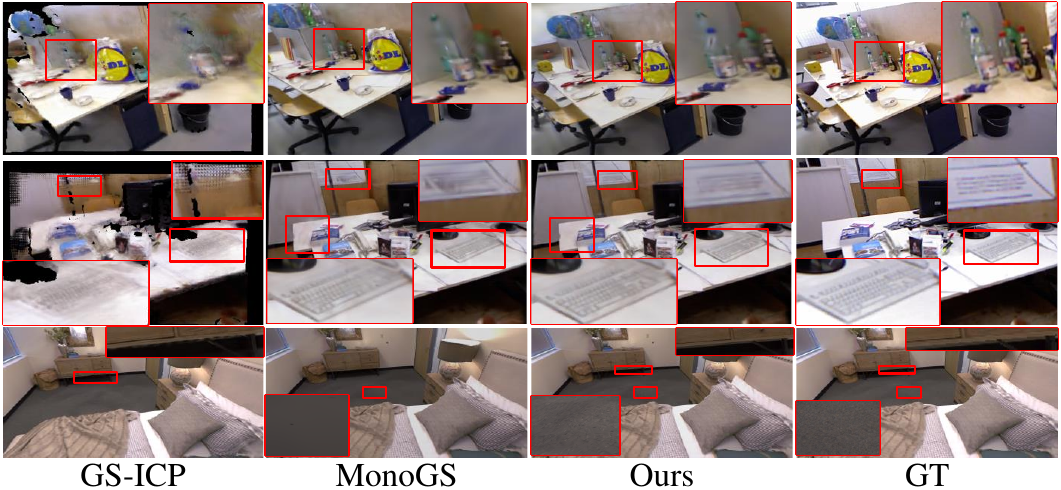}
    \caption{Testing results at 2$\times$ resolution on the Replica \cite{b25} and TUM datasets \cite{b26}. Other methods exhibit a decline in local regions, whereas our MipSLAM is capable of representing the detailed textures.}
    \label{f5}
\end{figure}

Tables \ref{table:replica} and \ref{table:tum} compare performance across multi-resolution settings on the Replica~\cite{b25} and TUM~\cite{b26} datasets. While all SOTA methods exhibit significant degradation (e.g., MonoGS~\cite{b5} declines from 39.51 dB to 24.01 dB in PSNR), our method remains robust. On Replica at 1/8 resolution, MipSLAM surpasses MonoGS~\cite{b5} by 5.46 dB and Scaffold-GS~\cite{b3} by 5.19 dB in PSNR. Across eight Replica sequences, MipSLAM achieves average gains of 5.85 dB over SplaTAM~\cite{b6}, 3.11 dB over MonoGS~\cite{b5}, and 4.12 dB over GS-ICP~\cite{b24}. On TUM, it outperforms SplaTAM, MonoGS, and GS-ICP by 1.26 dB, 0.40 dB, and 5.19 dB in PSNR, respectively. Moreover, while competing methods suffer widening performance gaps at lower resolutions on TUM (e.g., GS-ICP's SSIM drops from 0.851 at 4$\times$ to 0.646 at 1/4$\times$, a decline of 24.1\%), our method maintains an SSIM of 0.788 at 1/4$\times$. Notably, our online SLAM framework exceeds even offline reconstruction methods.

Fig. \ref{f4} demonstrates degradation at lower resolutions. Significant aliasing is observed on blinds in the first three rows. The fourth row exhibits blurred object edges and incomplete scene boundaries due to dilation in Vanilla 3DGS \cite{b1}, an effect mitigated by MipSLAM in the fifth row. Fig. \ref{f5} illustrates high-resolution artifacts: only MipSLAM accurately reconstructs bottle caps (first row); others exhibit blur in high-texture regions such as keyboards (second row); competing methods yield over-smoothed floors and missing table details (third row).

\subsection{Localization Accuracy}\label{abl:local}
As shown in Tab.~\ref{tab:tracking}, our MipSLAM achieves superior localization accuracy, outperforming all implicit and 3DGS-based methods with the lowest average ATE RMSE of 0.28 cm. Specifically, MipSLAM ranks first in 6 out of 8 scenes and achieves a 20\% relative improvement over the best implicit method GO-SLAM~\cite{b38} (0.28 vs.\ 0.35 cm). Compared to 3DGS-based approaches, while MonoGS~\cite{b5} and Photo-SLAM~\cite{b41} exhibit notable performance degradation on certain scenes (\emph{e.g.}, 2.06 cm on \textit{Off4} and 1.28 cm on \textit{Off2}, respectively), our method maintains consistently low error across all scenes with a standard deviation of only 0.065 cm, demonstrating its robustness in diverse indoor environments.

\section{CONCLUSION}

Traditional 3DGS tends to produce high-frequency artifacts under varying camera configurations, leading to degraded rendering quality. MipSLAM is an anti-aliased 3DGS-based SLAM system that co-optimizes rendering and pose estimation under a frequency-aware framework. Our contributions are: (1) the EAA algorithm, which uses adaptive elliptical sampling and Riemann summation to approximate pixel integrals; (2) SA-PGO, which models camera trajectories as spatiotemporal signals and applies spectral decomposition to the pose graph Laplacian for frequency-aware drift suppression. Experiments demonstrate that MipSLAM excels in cross-resolution rendering and pose estimation under significant resolution changes.

\printbibliography

\end{document}